\pdfoutput=1

\documentclass[11pt]{article}

\usepackage[utf8]{inputenc} 
\usepackage{times} 
\usepackage{latexsym} 
\usepackage[dvipsnames]{xcolor} 
\usepackage{amsmath} 
\usepackage{amssymb} 
\usepackage{wasysym} 
\usepackage{mathtools} 
\usepackage{annotate-equations} 
\usepackage{pifont} 
\usepackage{truncate} 
\usepackage{enumitem} 
\usepackage{csquotes} 
\usepackage{xurl} 
\usepackage[dont-mess-around]{fnpct} 
\usepackage{soul} 

\usepackage{ragged2e} 
\usepackage{float} 
\usepackage{dblfloatfix}

\usepackage{refcount}
\usepackage{footmisc}
\newcommand{\setfootnotemark}{%
  \refstepcounter{footnote}%
  \footnotemark[\value{footnote}]}

\usepackage{caption}
\usepackage{subcaption} 

\usepackage{graphicx} 

\usepackage{listings}
\definecolor{codegreen}{rgb}{0.0, 0.5, 0.0}
\definecolor{codegray}{rgb}{0.5, 0.5, 0.5}
\definecolor{codepurple}{rgb}{0.5, 0.0, 0.5}
\definecolor{backcolour}{rgb}{0.98, 0.98, 0.98}
\lstdefinestyle{codestyle}{
    backgroundcolor=\color{backcolour},   
    commentstyle=\color{codegreen},
    keywordstyle=\color{blue},
    numberstyle=\tiny\color{codegray},
    stringstyle=\color{codepurple},
    basicstyle=\ttfamily\footnotesize,
    breakatwhitespace=false,         
    breaklines=true,                 
    captionpos=b,                    
    keepspaces=true,                 
    numbers=left,                    
    numbersep=5pt,                  
    showspaces=false,                
    showstringspaces=false,
    showtabs=false,                  
    tabsize=2
}
\definecolor{delim}{RGB}{20,105,176}
\definecolor{numb}{RGB}{106, 109, 32}
\definecolor{string}{rgb}{0.64,0.08,0.08}

\lstdefinelanguage{json}{
    breaklines=true,
    postbreak=\raisebox{0ex}[0ex][0ex]{\ensuremath{\color{gray}\hookrightarrow\space}},
    breakatwhitespace=true,
    basicstyle=\ttfamily\small,
    upquote=true,
    morestring=[b]",
    stringstyle=\color{string},
    literate=
     *{0}{{{\color{numb}0}}}{1}
      {1}{{{\color{numb}1}}}{1}
      {2}{{{\color{numb}2}}}{1}
      {3}{{{\color{numb}3}}}{1}
      {4}{{{\color{numb}4}}}{1}
      {5}{{{\color{numb}5}}}{1}
      {6}{{{\color{numb}6}}}{1}
      {7}{{{\color{numb}7}}}{1}
      {8}{{{\color{numb}8}}}{1}
      {9}{{{\color{numb}9}}}{1}
      {\{}{{{\color{delim}{\{}}}}{1}
      {\}}{{{\color{delim}{\}}}}}{1}
      {[}{{{\color{delim}{[}}}}{1}
      {]}{{{\color{delim}{]}}}}{1},
}

\usepackage{array}
\usepackage{booktabs}
\usepackage{multirow}
\usepackage{tabularx}
\usepackage{tabulary}

\usepackage{tikz,pgfplots}
\pgfplotsset{compat=1.17}

\usepackage{xinttools} 
\usepackage{ifthen} 

\usepackage{lipsum}

\usepackage{todonotes}

\usepackage[T1]{fontenc}

\newcommand{\cmmnt}[1]{}

\newcommand{\cmark}{\ding{51}} 
\newcommand{\xmark}{\ding{55}} 
\newcommand{\partialmark}{$\ocircle$} 

\newcommand{\mainfig}{
    \begin{figure}[t]
        \centering
        \includegraphics[width=215px]{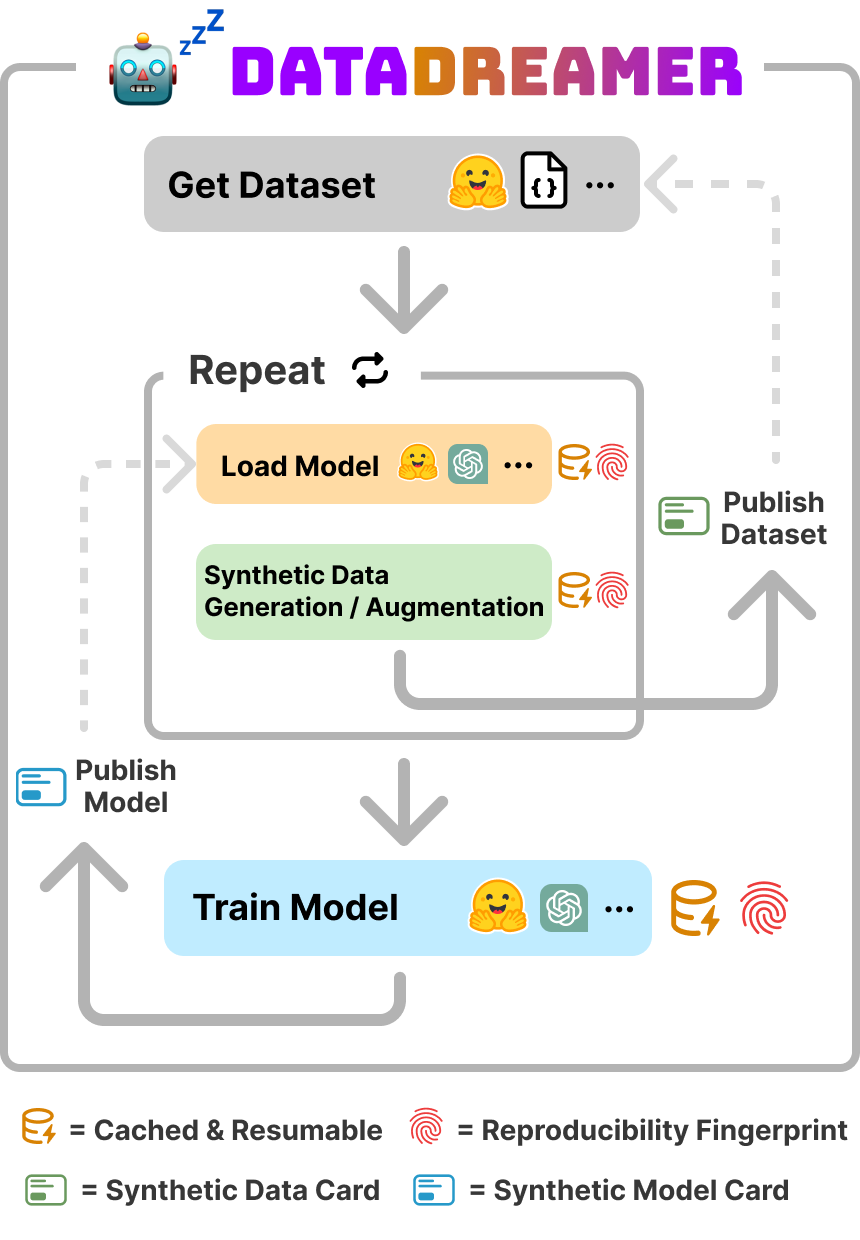}
        \caption{DataDreamer helps researchers implement many types of LLM workflows easier and makes reproducibility automatic and simple. These workflows often involve synthetic data generation with a LLM-in-the-loop and/or  fine-tuning, aligning, and distilling models.}
        \label{fig:main}
    \end{figure}
}

\newcommand{\logsfig}{
    \begin{figure*}[t]
        \centering
        \includegraphics[width=450px]{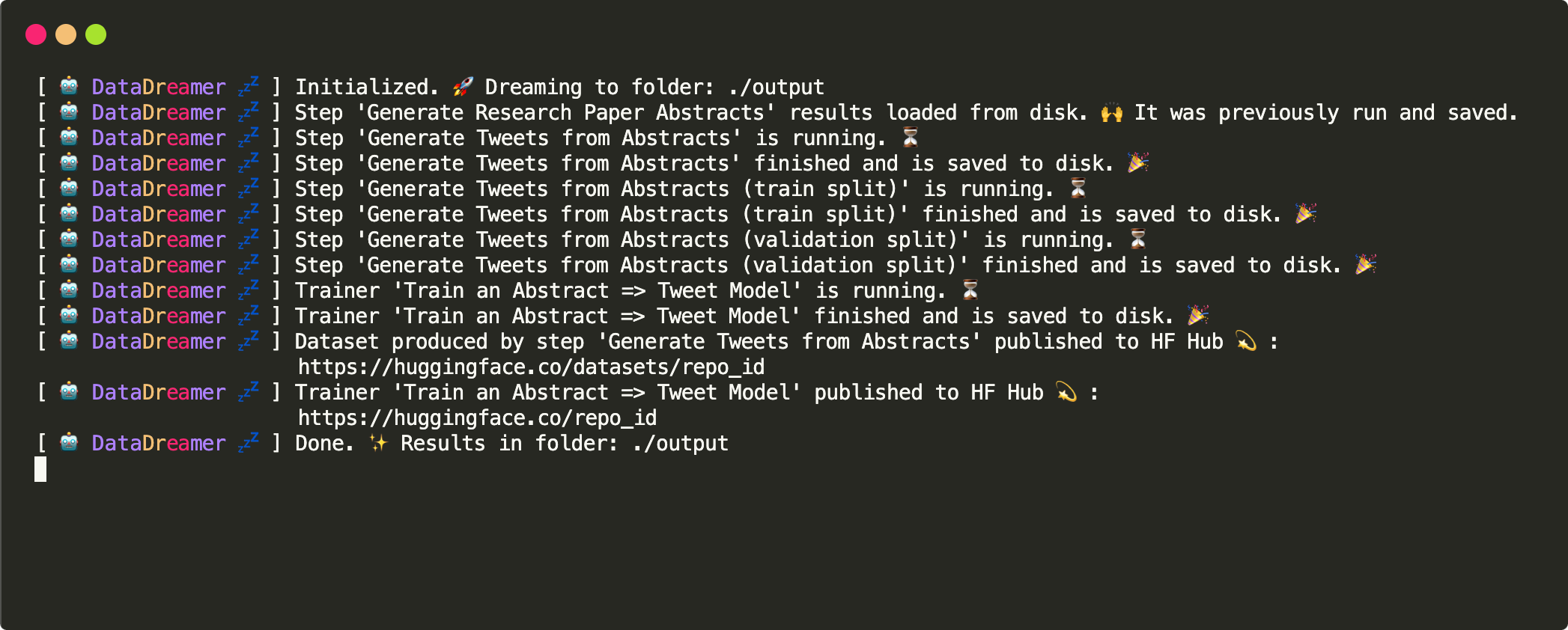}
        \caption{DataDreamer logs produced by the workflow in Example \ref{example:abstract_to_tweet} when resuming from a prior interrupted run.}
        \label{fig:logs}
    \end{figure*}
}

\newcommand{\featurematrixtable}{
    {
    \renewcommand{\arraystretch}{1}
    \begin{table*}[b!]
    \centering
    \small
    \setlength{\tabcolsep}{5pt}
    \begin{tabular}{lcccc}
    \toprule[\heavyrulewidth]
    \textbf{Feature} & \textbf{LangChain}\setfootnotemark\label{langchain}  & \textbf{Axlotl}\setfootnotemark\label{axlotl} & \textbf{HF Transformers + TRL}\setfootnotemark\label{hfeco} & \textbf{DataDreamer} \\ \midrule
    \multicolumn{5}{l}{\textit{Implementation}} \\ \midrule
    Accessible via Python API & \cmark & \xmark & \cmark & \cmark \\
    Built for Researchers & \xmark & \xmark & \cmark & \cmark \\
    \midrule
    \multicolumn{5}{l}{\textit{Integrations}} \\ \midrule
    Open Source Models & \cmark & \cmark & \cmark & \cmark \\
    Commercial \& API-based Models & \cmark & \xmark & \xmark & \cmark \\ \midrule
    \multicolumn{5}{l}{\textit{Tasks}} \\ \midrule
    Prompting \& Prompt ``Chaining'' & \cmark & \xmark & \xmark & \cmark \\
    Synthetic Data Generation \& Augmentation & \cmark & \xmark & \xmark & \cmark \\
    Fine-tuning LLMs & \xmark & \cmark & \cmark & \cmark \\ 
    Instruction-tuning LLMs & \xmark & \cmark & \cmark & \cmark \\
    Aligning LLMs & \xmark & \cmark & \cmark & \cmark \\
    Training Classifier Models & \xmark & \xmark & \cmark & \cmark \\
    Training Embedding Models & \xmark & \xmark & \xmark & \cmark \\ \midrule
    \multicolumn{5}{l}{\textit{Conveniences}} \\ \midrule
    Caching & \partialmark & \xmark & \xmark & \cmark \\
    Resumability & \xmark & \cmark & \partialmark & \cmark \\
    Simplifies Boilerplate Code (tokenization, etc.) & \cmark & \cmark & \xmark & \cmark \\
    Simplifies Multi-GPU Inference and Training & \xmark & \partialmark & \xmark & \cmark \\
    Publishing Datasets \& Models & \xmark & \partialmark & \cmark & \cmark \\ \midrule
    \multicolumn{5}{l}{\textit{Open Science and Reproducibility}} \\ \midrule
    Reproducibility Fingerprints & \xmark & \xmark & \xmark & \cmark \\
    Saves Intermediate Outputs & \xmark & \xmark & \xmark & \cmark \\
    Synthetic Data and Model Cards & \xmark & \xmark & \xmark & \cmark \\
    \bottomrule[\heavyrulewidth]
    \end{tabular}
    \caption{We compare feature coverage between other popular libraries and solutions available to researchers today that target similar workflows. DataDreamer integrates these features into a single library with a standardized interface making experimentation and chaining data between tasks simple. (\xmark = No; \cmark = Yes; \partialmark ~= Partial Support) 
    }
    \label{table:featurematrixtable}
    \end{table*}
    }
}
\newcommand{\builtinstable}{
    {
    \renewcommand{\arraystretch}{1.5}
    \renewcommand\tabularxcolumn[1]{m{##1}}%
    \begin{table*}[b!]
    \centering
    \small
    \setlength{\tabcolsep}{5pt}
    \begin{tabularx}{\textwidth}{llX}
    \toprule[\heavyrulewidth]
    \textbf{Type} & & \textbf{Examples} \\ \midrule
    \multirow{3}*{\textbf{Steps}} & Load a Dataset & \texttt{DataSource}, \texttt{HFHubDataSource}, \texttt{JSONDataSource}, \texttt{CSVDataSource}, ...  \\
    & Prompting & \RaggedRight \texttt{Prompt}, \texttt{RAGPrompt}, \texttt{ProcessWithPrompt}, \texttt{FewShotPrompt}, \texttt{DataFromPrompt}, \texttt{DataFromAttributedPrompt}, \texttt{FilterWithPrompt}, \texttt{RankWithPrompt}, \texttt{JudgeGenerationPairsWithPrompt}, ... \\
   & Other & \texttt{Embed}, \texttt{Retrieve}, \texttt{CosineSimilarity}, ... \\
    \textbf{Models} & & \RaggedRight \texttt{OpenAI}, \texttt{OpenAIAssistant}, \texttt{HFTransformers}, \texttt{CTransformers}, \texttt{VLLM}, \texttt{Petals}, \texttt{HFAPIEndpoint}, \texttt{Together}, \texttt{MistralAI}, \texttt{Anthropic}, \texttt{Cohere}, \texttt{AI21}, \texttt{Bedrock}, \texttt{Vertex}, ... \\
    \textbf{Trainers} & & \RaggedRight \texttt{TrainOpenAIFineTune}, \texttt{TrainHFClassifier}, \texttt{TrainHFFineTune}, \texttt{TrainSentenceTransformer}, \texttt{TrainHFDPO}, \texttt{TrainHFPPO}, ... \\
    \bottomrule[\heavyrulewidth]
    \end{tabularx}
    \caption{A few examples of built-in steps, models, and trainers available in DataDreamer.}
    \label{table:builtins}
    \end{table*}
    }
}
\newcommand{\syntheticcardtable}{
    \begin{table}[t]
    \centering
    \small
    \begin{tabular}{p{0.94\linewidth}}
    \toprule[\heavyrulewidth]
    \multicolumn{1}{l}{\textbf{Date \& Time}} \\ \midrule
    The date and time the step or trainer was run. This is important to document when using API-based LLMs that can be updated over time. \\ \midrule
    \multicolumn{1}{l}{\textbf{Dataset Name \& Card}} \\ \midrule
    The name of any datasets used as part of a step or trainer's operation along with their data cards. \\ \midrule
    \multicolumn{1}{l}{\textbf{Model Name \& Card}} \\ \midrule
    The name of any models used in a step or trainer's operation along with their model cards. \\ \midrule
    \multicolumn{1}{l}{\textbf{URL}} \\ \midrule
    A URL that can be referenced for more information about the step or trainer. \\ \midrule
    \multicolumn{1}{l}{\textbf{License}} \\ \midrule
    Any known license that may apply as a result of a model or dataset being used in a step or trainer. \\ \midrule
    \multicolumn{1}{l}{\textbf{Citations}} \\ \midrule
    Citations for datasets and models used in a trainer. \\ \midrule
    \multicolumn{1}{l}{\textbf{Reproducibility Fingerprint}} \\ \midrule
    A hash of all inputs, arguments, and configurations that may affect reproducibility for a step or trainer. When steps and trainers are chained in a multi-stage workflow, the reproducibility hash is computed recursively through the chain. These fingerprints can be used to compare if two workflows within DataDreamer are exactly identical. \\ \midrule
    \multicolumn{1}{l}{\textbf{Other Reproducibility Information}} \\ \midrule
    Other miscellaneous reproducibility information such as environment information, system information, and versions of packages and dependencies. \\
    \bottomrule[\heavyrulewidth]
    \end{tabular}
    \caption{Information automatically recorded in a synthetic data card or synthetic model card. An example synthetic data card can found in Appendix \ref{sec:syntheticdatacard}.}
    \label{table:syntheticcardtable}
    \end{table}
}
\newcommand{\abstractotweetexample}{
\lstinputlisting[label={example:abstract_to_tweet},language=Python, caption={In this demonstration snippet, DataDreamer generates a fully synthetic dataset of tweets summarizing research paper abstracts and then trains a smaller T5 distilled model \citep{t5} to perform the task and publishes both the synthetic dataset and the trained model. DataDreamer makes it simple to chain data from each step in the workflow to the next and automatically caches each step of this workflow to the \texttt{./output/} folder to allow interruption and resumability at any point in the script. The standardized API also makes it easy to switch to and experiment with different models, both open source and commercial, for generation and training.},float=*]{resources/examples/abstract_to_tweet.py}
}
\newcommand{\instructexample}{
\lstinputlisting[label={example:instruct},language=Python, caption={In this demonstration snippet, we instruction-tune a model \citep{instructgpt,tinyllama,alpaca}. DataDreamer reduces boilerplate around tokenization, caching, training resumability, multi-GPU training, parameter-efficient fine-tuning, and more.}]{resources/examples/instruct.py}
}
\newcommand{\alignexample}{
\lstinputlisting[label={example:align},language=Python, caption={In this demonstration snippet, we align a model using DPO \citep{dpo,tinyllama,openorca,orca}. DataDreamer reduces boilerplate around tokenization, caching, training resumability, multi-GPU training, parameter-efficient fine-tuning, and more.}]{resources/examples/align.py}
}
\newcommand{\selfrewardingexample}{
\lstinputlisting[label={example:selfrewarding},language=Python, caption={This demonstration snippet implements a simplified version of the self-rewarding LLMs \citep{selfrewarding} procedure. This workflow involves using an LLM to judge its own generations in order to self-align and self-improve itself over a number of rounds. DataDreamer allows this complex multi-stage workflow to be implemented intuitively, without needing to split generation and training logic into separate files and without needing to involve a launcher like \texttt{torchrun} to perform multi-GPU training. DataDreamer also makes this complex multi-round, multi-stage workflow automatically cachable and resumable.}]{resources/examples/selfrewarding.py}
}
\newcommand{\augmentexample}{
\lstinputlisting[label={example:augment},language=Python, caption={In this demonstration snippet, we augment an existing dataset, HotpotQA \citep{hotpotqa}, a multi-hop QA dataset. DataDreamer makes it easy to perform synthetic dataset augmentation with a LLM. In this example, we add intermediate questions required to solve the multi-hop question.}]{resources/examples/augment.py}
}
\newcommand{\syntheticdatacardexample}{
\lstinputlisting[label={example:syntheticdatacard},language=json, caption={A JSON representation of an example automatically generated synthetic data card produced by DataDreamer for Example \ref{example:abstract_to_tweet}. Synthetic data cards and model cards are automatically produced by recursively tracing through any steps, models, and trainers used to produce a given dataset or model. Each step, model, and trainer has associated metadata such as license information and citation information. DataDreamer collects this information and produces a synthetic data card (or model card) that reports the information along with reproducibility information like the reproducibility fingerprint.}]{resources/examples/synthetic_data_card.json}
}

\usepackage[final]{acl}

\usepackage{times}
\usepackage{latexsym}

\usepackage[T1]{fontenc}

\usepackage[utf8]{inputenc}

\usepackage{microtype}

\usepackage{inconsolata}

\title{DataDreamer: A Tool for Synthetic Data Generation and\\ Reproducible LLM Workflows}

\author{Ajay Patel \\
  University of Pennsylvania \\
  \texttt{ajayp@upenn.edu} \\\And
  Colin Raffel \\
  University of Toronto \\
  Vector Institute \\
  \texttt{craffel@gmail.com} \\\And
  Chris Callison-Burch \\
  University of Pennsylvania \\
  \texttt{ccb@upenn.edu}}

\begin{document}
\maketitle

\begin{abstract}
Large language models (LLMs) have become a dominant and important tool for NLP researchers in a wide range of tasks. Today, many researchers use LLMs in synthetic data generation, task evaluation, fine-tuning, distillation, and other model-in-the-loop research workflows. However, challenges arise when using these models that stem from their scale, their closed source nature, and the lack of standardized tooling for these new and emerging workflows. The rapid rise to prominence of these models and these unique challenges has had immediate adverse impacts on open science and on the reproducibility of work that uses them. In this ACL 2024 theme track paper, we introduce DataDreamer, an open source Python library that allows researchers to write simple code to implement powerful LLM workflows. DataDreamer also helps researchers adhere to best practices that we propose to encourage open science and reproducibility. The library and documentation are available at: \url{https://github.com/datadreamer-dev/DataDreamer}.
\end{abstract}
\section{Introduction}
\label{sec:introduction}

\mainfig

While large language models (LLMs) have established a new era in NLP research 
through the prompt-and-predict paradigm that has proven effective on a wide variety of tasks, the use of these models has come with significant drawbacks \citep{promptsurvey}. Many popular models like GPT-4 \citep{gpt4} are closed source and behind a remote API, while running models locally can be technically complex and expensive due to their scale. Moreover, the now well-established prompting paradigm can be brittle with results widely varying between different models, configurations, and environments \citep{promptsensitivity,compressingtruth}. These challenges have made it difficult for researchers to share, reproduce, extend, and compare work, hindering the rate of research progress.

\featurematrixtable
\vspace*{0.5em}
In context of the rapid shift to using these large models in research, we introduce DataDreamer, our open source Python package that provides both practical utility to researchers and scientific utility to the community:

\vspace*{0.5em}

\begin{itemize}
\item DataDreamer helps researchers implement state-of-the-art emerging workflows involving LLMs such as synthetic data generation, fine-tuning, instruction-tuning, and alignment.  It simplifies implementations by providing a single library with a standardized interface for many of these tasks while reducing technical complexity around switching between models, caching, resumability, logging, multi-GPU inference and training, using adapter and quantization optimizations, and publishing open datasets and models. 

\item DataDreamer makes chaining data between tasks, an increasingly common practice, simple. For example, a user can generate data with a synthetic data workflow and then fine-tune on that synthetic data.

\item DataDreamer helps researchers implement workflows while crucially producing output that is compatible with open science and reproducible ideals with minimal effort, through automatic caching, reproducibility fingerprints, and more best-practice artifacts.
\end{itemize}
\pagebreak
\section{LLM Workflows}
\label{sec:emerging-llm-workflows}

To motivate DataDreamer, we first discuss the LLM workflows that it supports. We discuss challenges to open science that arise from these usage patterns. In this paper, we do not seek to validate or critique these approaches. Instead, we offer a solution to implement them and make them reproducible. These LLM workflows are often used in combination with each other \citep{selfrewarding}, and orchestration of multi-stage workflows is frequently implemented through multiple shell or Python scripts. Reproducing these multi-stage workflows is challenging as shell scripts may rely upon a particular author's job scheduler or environment and require execution in a specific order. In Section \ref{sec:datadreamer} and \ref{sec:reproducibility}, we discuss how DataDreamer's task orchestration, caching system, and simple multi-GPU training make it easier to implement these multi-stage workflows in a single Python program, minimizing these issues.

\paragraph{Synthetic Data Generation} Recent work has explored using LLMs to create synthetic data for tasks or to augment existing datasets to boost task performance \citep[][\textit{inter alia}]{attrprompt,syntheticdata1,syntheticdata2,syntheticdata3,syntheticdata4,syntheticdata5,syntheticdata7}. Synthetic data generation involves using a LLM once or multiple times in a multi-stage workflow to process data, sometimes referred to as ``chaining'' \citep{minichain}. When prompting LLMs to generate or augment datasets, a reproducibility challenge that arises is ``prompt sensitivity'' where even small variations in a prompt can lead to significantly different results \citep{promptsensitivity}. Moreover, it is imperative to tag synthetically generated datasets because of model degradation concerns \citep{recursivetraining}.

\paragraph{LLMs for Task Evaluation} Another increasingly common workflow is using LLMs as judges or as automatic metrics for evaluating a model's performance on a task \citep[][\textit{inter alia}]{llmjudge1,llmjudge2,llmjudge3,llmjudge4}. Many of the reproducibility challenges applicable to synthetic data also arise here.

\paragraph{Fine-tuning and Alignment} Another common workflow is the creation of task-specific expert models using knowledge from larger models to create smaller, more efficient models via fine-tuning and distillation \citep{finetuning1,finetuning2,finetuning3}. Instruction-tuning is fine-tuning that allows base pre-trained models to better follow natural language human instruction and improve their generalized task performance \citep{instructgpt,flan,t0,natinstructions}. Closely related, alignment techniques steer model responses towards those more preferable to humans \citep{learningtosummarize,constitutionalai,dpo}. Implementing resumability and efficient training techniques are practical challenges often faced. Reproducibility challenges include sharing exact data and hyperparameters.

\paragraph{Self-improving LLMs} Self-improving LLMs through self-feedback training loops is an increasingly active area of research interest \citep{selfimprove,selfinstruct,gvconsistency,selfplay,selfrewarding,phi}. These workflows can be uniquely complex to both implement and reproduce due to requiring multiple rounds that chain together synthetic data generation, automatic evaluation, and model re-training. DataDreamer supports all of these workflows and makes it simple to chain data between them.

\footnotetext[\getrefnumber{langchain}]{\url{https://github.com/langchain-ai/langchain}}
\footnotetext[\getrefnumber{axlotl}]{\url{https://github.com/OpenAccess-AI-Collective/axolotl}}

\pagebreak

\footnotetext[\getrefnumber{hfeco}]{\citet{transformers,trl}}

\section{Demonstration and Examples}
\label{sec:examples}

Before delving into the structure and implementation of DataDreamer, we first provide a simple demonstration of DataDreamer's capabilities and API through an example synthetic data generation and distillation workflow in Example \ref{example:abstract_to_tweet}. The LLM used in this example is GPT-4 \citep{gpt4}. As an initial step, the example uses the LLM to generate 1,000 NLP research paper abstracts. The LLM is then used to summarize those abstracts in a tweet-like style. These two steps result in a fully synthetic dataset of abstracts and tweets summarizing them. Using a trainer, this synthetic dataset is then distilled to a small, local model that is capable of summarizing paper abstracts in a tweet-like style. As a final step, the example demonstrates how both the synthetic dataset and the trained model can be published and shared. For illustrative purposes, we demonstrate a sample generation of the trained model's output on this paper's abstract:
\\[.5em]\centerline{\fbox{\begin{minipage}{17.5em}
\begin{flushleft}\small{
``Introducing DataDreamer, an open source Python library for advanced \#NLP workflows. It offers easy code to create powerful LLM workflows, addressing challenges in scale, closed source nature, and tooling. A step towards open science and reproducibility! \#AI \#MachineLearning''
}\end{flushleft}\end{minipage}}}\\[.5em]

Further example workflows can be found in the Appendix (Example \ref{example:instruct}, Example \ref{example:align}, Example \ref{example:selfrewarding}, Example \ref{example:augment}).
\abstractotweetexample

\section{DataDreamer}
\label{sec:datadreamer}

DataDreamer is an open source Python package that allows researchers to implement all of the LLM workflows discussed in Section \ref{sec:emerging-llm-workflows} using a single library. DataDreamer provides a standardized interface for prompting and training models, abstracting away vendor-specific libraries and tooling. This makes research code simpler to implement, modify, experiment with, and share with others. DataDreamer integrates with other open source LLM libraries like \texttt{transformers} \citep{transformers} and \texttt{trl} \citep{trl}, as well as commercial model APIs like OpenAI and Anthropic\footnote{\url{https://www.anthropic.com/}} for commercial LLMs \citep{gpt3}. Moreover, DataDreamer automatically implements the best practices for reproducibility discussed in Section \ref{sec:reproducibility}.

\builtinstable

\subsection{Installation}
DataDreamer can be installed with:
\begin{lstlisting}[language=bash,firstnumber=1,basicstyle=\ttfamily\normalsize,numbers=none]]
pip install datadreamer.dev
\end{lstlisting}

\subsection{Sessions}

All code using the DataDreamer library is placed within a ``session'' using a Python context manager instantiated using the \texttt{with} keyword:

\begin{lstlisting}[language=bash,firstnumber=1,basicstyle=\ttfamily\small,numbers=none]]
from datadreamer import DataDreamer

with DataDreamer("./output"):
    ...
\end{lstlisting}

Workflow tasks can be run within the session context manager. These tasks are called ``steps'' (loading a dataset, prompting a model, etc.) or ``trainers''. The session allows DataDreamer to automatically organize the resulting datasets, outputs, caches, training checkpoints, and trained models that  result from tasks run within the session into the \texttt{./output/} folder. Each step in a workflow assigns a custom descriptive name for its subfolder under \texttt{./output/}. DataDreamer sessions automatically provide user-friendly logging around workflow tasks run within the session (see Figure \ref{fig:logs}).

\subsection{Steps}

Steps are the core operators in a DataDreamer session. A step in DataDreamer transforms from an input dataset to an output dataset \citep{datasets}. This is useful for tasks like generating synthetic data from LLMs, or data augmentation for existing datasets. The output of one step can be directly used as the input to another step or as the input to a trainer, allowing users to chain together multiple steps/trainers to create complex workflows. DataDreamer comes with a number of built-in steps for common operations in LLM workflows, some examples of which can be seen in Table \ref{table:builtins}. Useful standard data processing operations such as \texttt{.map()}, \texttt{.filter()}, and  \texttt{.shuffle()} can also quickly be applied to the output of a step for custom processing. DataDreamer uses memory-mapping to handle large datasets stored on disk and can be run lazily over iterable, streaming datasets.

\subsection{Models}

Models can be loaded in a DataDreamer session and then be passed as an argument to steps like \texttt{FewShotPrompt} and \texttt{ProcessWithPrompt}. DataDreamer creates a standardized interface for accessing open source and commercial LLMs. It includes interfaces for embedding models as well as LLMs. Examples of supported models and model providers can be found in Table \ref{table:builtins}.

\subsection{Trainers}

Trainers can train on a dataset produced by a step in a DataDreamer workflow. The dataset may be loaded from an external source or produced as the output of a step in a multi-step workflow. DataDreamer's trainers support a wide variety of techniques and tasks including fine-tuning, instruction-tuning, alignment via RLHF \cite{instructgpt} and DPO \cite{dpo}, distillation, training classifiers, and training embedding models. Examples of supported techniques are shown in Table \ref{table:builtins}.

\subsection{Caching and Sharing Workflows}

Caching has practical utility in LLM workflows as these large models can be both computationally and financially expensive to run. Therefore, eliminating re-computation can save both time and resources.  Caching in DataDreamer happens at multiple levels. When a step or trainer is completed, its resulting dataset or trained model is saved to disk and loaded from disk if the step or trainer is executed again with the same inputs and arguments, instead of being run again. Additionally, DataDreamer caches at the model-level, caching the results of prompts or texts being run against a model to a SQLite database file. During training, DataDreamer similarly automatically saves checkpoints and resumes from them if interrupted and restarted. Caching uses minimal disk space (storing mainly text) and adds minimal overhead in these workloads dominated by heavy model inference computation, but can be granularly disabled if desired.

DataDreamer's cache system allows a researcher to share both their workflow script and their session output folder with others, giving them access to useful caches and saved outputs. These allow others to easily reproduce and extend the entire workflow while also benefiting from avoiding expensive computations when unnecessary. For example, a researcher could extend another researcher's workflow by adding another step at the end. Only the additional added step would need to be computed, while all of the original steps could have their results loaded from disk.

\subsection{Resumability}

Caching allows resumability during development, so scripts can be interrupted and resumed. This allows graceful handling of crashes, server preemption, and other situations where only a portion of a workflow was previously computed. Furthermore, caching can be useful during experimentation of a workflow. For example, when modifying a single prompt in the middle of a multi-step synthetic data generation workflow, the change may only affect a certain number of inputs to the next step. If so, only that portion of the work will be re-computed.

\subsection{Sharing Open Data and Open Models}

DataDreamer provides convenient utilities for exporting and publishing datasets and trained models produced by steps or trainers. Resources can be exported to disk or published to the Hugging Face Hub \footnote{https://huggingface.co/}. When resources are published, DataDreamer can automatically upload a demonstration snippet and set up the live demonstration widget on the Hugging Face Hub, which makes shared resources easily usable. Additionally, these resources are automatically given appropriate metadata such as tags clearly indicating when data is synthetically generated and its source LLM.  DataDreamer also produces what we call ``synthetic data cards'' and ``synthetic model cards''. Synthetic data and model cards are automatically produced by recursively tracing through all steps, models, and trainers that DataDreamer used to produce the dataset or model. Each step, model, and trainer has associated metadata including license information and citation information. DataDreamer collects this information and produces a synthetic data card (or model card) that reports the information along with reproducibility information for each step, model, and trainer in the workflow. The information collected in our cards is defined in Table \ref{table:syntheticcardtable}.

\syntheticcardtable

These automatically generated synthetic data cards and model cards can aid in preventing contamination of pre-training sources with model-generated synthetic data. As synthetic data generation becomes more prevalent, contamination can be a concern due to the performance degradation that has been observed when synthetic datasets are shared and trained on, possibly without the knowledge of the model developer \citep{recursivetraining}. DataDreamer's cards can also help other researchers understand what license restrictions may apply to the synthetically generated data, among other usability concerns. These automatically generated cards are not a replacement for traditional data cards and model cards \citep{datacard,modelcard} that recommend a wider set of important attributes such as potential dataset biases. Instead, they provide supplemental information that is crucial to the usability and reproducibility of LLM workflows. We encourage researchers to review and add information that cannot be automatically detected to our generated cards.

\subsection{Efficiency and Optimizations}

LLMs workflows often benefit from or require certain optimizations to be applied in order to load or process the scale of data and models typically used. DataDreamer supports many of the common optimizations that researchers may want to apply.

\paragraph{Parallelization} DataDreamer supports running steps in background processes and running steps concurrently to easily implement parallel task orchestration in a workflow.

\paragraph{Quantization and Adapters} DataDreamer supports quantization of model weights that can reduce memory usage \citep{quantization} as well as parameter-efficient fine-tuning techniques like LoRA adapters \citep{lora,peft}. It standardizes using these optimizations across different model architectures and minimizes boilerplate, making it as simple as a single argument to configure training with LoRA in Example \ref{example:abstract_to_tweet}. DataDreamer attempts to create uniform support for features across all of its supported integrations when possible. So while the underlying \texttt{sentence\_transformers} and \texttt{transformers} libraries do not support training embedding models with LoRA \citep{sentence_transformers,transformers}, DataDreamer supports this, which extends the benefits of LoRA to these models.

\paragraph{Multi-GPU Usage} DataDreamer makes it simple to load models on multiple GPUs and train models on multiple GPUs with PyTorch FSDP \citep{pytorch,fsdp}. For example, training a model on multiple GPUs is as simple as passing a list of \texttt{torch.device}s to the \texttt{device} parameter of a trainer (\texttt{device=["cuda:0", "cuda:1"]}). DataDreamer automatically configures FSDP and launches distributed processes within the session so that a command line launcher like \texttt{torchrun} never has to be used, simplifying multi-GPU training. The use of \texttt{torchrun} can often force complex, multi-stage workflows being split into multiple scripts launched via shell scripts since training portions need to be isolated from data generation or data processing portions. This added complexity in running the workflow end-to-end can make reproducibility challenging. With DataDreamer, workflows do not need to be re-orchestrated around portions needing to be launched via \texttt{torchrun}. Since DataDreamer handles this distributed orchestration automatically, users can build multi-stage workflows involving data generation, data processing, and training on multiple GPUs all in a single Python program, obviating the use of orchestration through multiple shell scripts. Example \ref{example:selfrewarding} in the Appendix provides an example of such a workflow.

\subsection{Configuration and Extensibility}

DataDreamer seeks to minimize configuration and boilerplate code that for most research workflows do not need to be customized, for example automatically handling tokenization and applying the correct padding, among other tasks. DataDreamer applies sensible defaults and standard research practices to minimize configuration. Some researchers, however, may need to customize these choices and the option to override and extend is provided and well-documented.


\logsfig

\section{Reproducibility}
\label{sec:reproducibility}

We outline a few best practices, specific to the emerging use of LLMs in research workflows that DataDreamer adopts. We believe instituting these practices can alleviate a number of reproducibility concerns. Of course, when closed-source models are involved, these concerns can never be fully eliminated (see Section \ref{sec:limitations} for further discussion on limitations). We discuss how DataDreamer makes it easier to implement these practices or automatically implements these practices in this section.

\paragraph{Adaptable to Model Substitution} While experimental workflows can often be sensitive to model choice and the transferability of prompts can be unreliable \citep{promptsurvey}, for reproducibility purposes and for ease of experimentation, workflow implementation code should attempt to minimize dependence on a specific model and should allow other researchers to easily substitute one LLM for another. This can also be useful if a model is not accessible to another researcher or if a model has become obsolete. DataDreamer's API and model abstractions make model substitution simple.

\paragraph{Sharing Prompts} Exact prompts used should be shared since even minor variations can significantly impact performance \citep{promptsensitivity}. DataDreamer makes it easy to share an entire workflow and session output folder. DataDreamer can also help ensure a re-implementation is exactly identical between two experimental setups by comparing the reproducibility fingerprints of individual steps or the entire workflow in aggregate.

\paragraph{Sharing Intermediate Outputs} In multi-stage workflows, intermediate outputs should be shared for inspection and analysis by other researchers as well as for extendability purposes. DataDreamer makes this simple by automatically saving the results of each step in a multi-stage workflow in an easily inspectable Hugging Face \texttt{datasets} format \citep{datasets}. When API-based LLMs are used, there is greater risk to reproducibility. DataDreamer allows workflows to be exactly reproduced from caches in the session output folder, even if the remote API is no longer available.

\paragraph{Synthetic Data Cards and Model Cards} Synthetic data and model cards can help other researchers understand the source of synthetic data, license restrictions that may apply, citations that may apply, among other attributes. Importantly, these cards and other metadata-like tags can help prevent contamination of pre-training data \citep{recursivetraining}. Finally, these cards carry reproducibility information, useful for validating two experimental setups as identical.

\paragraph{Sharing Optimization Configurations} Optimizations like quantization can have an effect on generations \citep{compressingtruth}. DataDreamer's reproducibility fingerprints account for these configurations and with its easily shareable workflows, DataDreamer makes it easy to reproduce an exact workflow, along with configured optimizations.

\paragraph{Environment-Agnostic Code} For reproducibility, code should attempt to minimize dependence on local environments, job schedulers, shell scripts, etc. DataDreamer helps make this easier by providing tools for workflow orchestration (steps, parallelization, managed distributed processes for multi-GPU training) that can be all be done within Python. DataDreamer also minimizes dependencies on local file paths, by organizing results and outputs into the session output folder automatically.
\section{Conclusion}
\label{sec:conclusion}

The current moment in NLP research and recent progress is exciting yet raises important questions for the community. We introduce DataDreamer, an open source Python package for implementing common patterns and workflows involving LLMs. We believe DataDreamer provides both practical and scientific utility to the research community and that its adoption can help advance the rate of research progress in workflows involving LLMs by making implementation easier and making research output reproducible and extendable. 
\section * {Limitations}
\label{sec:limitations}

In this work, we outline best practices and implement these practices in an open source system called DataDreamer. We believe these contributions can help aid open science in our field, however, we acknowledge that as long as the research community chooses to use closed-source models for experiments, especially those served behind an API on remote servers, challenges to reproducibility are inevitable. With DataDreamer, we provide a way to reproduce and further analyze some of these experiments long after these remote APIs may be changed or unavailable through the session-based caching system as well as provide a way to easily substitute models where needed through abstractions. To the best of our knowledge, there are no significant ethical considerations that arise from this work. We believe the broader impacts of this work to be largely positive, making state-of-the-art LLM workflows both easier and more accessible to implement and reproduce as well as reducing carbon emissions through DataDreamer's caching system that helps researchers avoid expensive re-computation when possible.
\section*{Acknowledgements}

This research is supported in part by the Office of the Director of National Intelligence (ODNI), Intelligence Advanced Research Projects Activity (IARPA), via the HIATUS Program contract \#2022-22072200005. The views and conclusions contained herein are those of the authors and should not be interpreted as necessarily representing the official policies, either expressed or implied, of ODNI, IARPA, or the U.S. Government. The U.S. Government is authorized to reproduce and distribute reprints for governmental purposes notwithstanding any copyright annotation therein.

\bibliography{custom}

\appendix\onecolumn
\section{Instruction-Tuning a LLM}
\label{sec:instructexample}

\instructexample
\newpage
\section{Aligning a LLM}
\label{sec:alignexample}

\alignexample
\newpage
\section{Self-Rewarding LLMs}
\label{sec:selfrewardingexample}

\selfrewardingexample
\newpage
\section{Augmenting an Existing Dataset}
\label{sec:augmentexample}

\augmentexample
\newpage
\section{Example Synthetic Data Card}
\label{sec:syntheticdatacard}

\syntheticdatacardexample
\newpage

\end{document}